\documentclass[11pt,a4paper]{article}
\usepackage{times,latexsym}
\usepackage{url}
\usepackage[T1]{fontenc}

\usepackage[acceptedWithA]{tacl2018v2}
\usepackage{enumitem}

\usepackage{pgfplots}
\usepgfplotslibrary{groupplots}
\usetikzlibrary{patterns}
\pgfplotsset{every tick label/.append style={font=\footnotesize}}
\pgfplotsset{compat = newest}

\usepackage{ulem}

\definecolor{C1}{HTML}{1E88E5}
\definecolor{C2}{HTML}{D81B60}
\definecolor{C3}{HTML}{FFC107}
\definecolor{C4}{HTML}{004D40}
\definecolor{C5}{HTML}{D55E00}
\definecolor{C6}{HTML}{785EF0}

\usepackage{xspace,mfirstuc,tabulary}

\newif\iftaclinstructions
\taclinstructionsfalse 
\iftaclinstructions
\renewcommand{\confidential}{}
\renewcommand{\anonsubtext}{(No author info supplied here, for consistency with
TACL-submission anonymization requirements)}
\newcommand{\instr}
\fi

\iftaclpubformat 

\else

\fi

\usepackage[utf8]{inputenc} 
\usepackage[T1]{fontenc}    
\usepackage{hyperref}       
\usepackage{url}            
\usepackage{booktabs, subcaption}       
\usepackage{amsfonts}       
\usepackage{nicefrac}       
\usepackage{microtype}      
\usepackage{xcolor}         
\usepackage{graphicx}

\usepackage{multicol}
\usepackage{subcaption}

\usepackage{cleveref}
\crefname{section}{\S}{\S\S}
\Crefname{section}{\S}{\S\S}
\crefname{table}{Table}{Tables}
\crefname{figure}{Figure}{Figures}
\crefname{algorithm}{Algorithm}{}
\crefname{equation}{eq.}{}
\crefname{appendix}{Appendix}{}
\crefformat{section}{\S#2#1#3}

\usepackage{color, colortbl} 
\definecolor{g}{gray}{0.90}
\definecolor{pistachio}{rgb}{0.58, 0.77, 0.45}
\definecolor{persianred}{rgb}{0.8, 0.2, 0.2}
\definecolor{rojo}{rgb}{1, 0.45, 046}

\usepackage{todonotes}
\makeatletter
\newcommand*\iftodonotes{\if@todonotes@disabled\expandafter\@secondoftwo\else\expandafter\@firstoftwo\fi} 
\makeatother

\usepackage{xspace}

\newcommand{\Ma}{CNN$_{\textrm{\footnotesize{NoPre}}}$\xspace}
\newcommand{\Mb}{RN$_{\textrm{\footnotesize{NoPre}}}$\xspace}
\newcommand{\Mc}{RN$_{\textrm{\footnotesize{IMG}}}$\xspace}
\newcommand{\Md}{RN$_{\textrm{\footnotesize{CLIP}}}$\xspace} 
\newcommand{\Me}{ViT$_{\textrm{\footnotesize{CLIP}}}$\xspace}

\title{Unit Testing for Concepts in Neural Networks}

\author{%
  Charles Lovering \\
  Department of Computer Science\\
  Brown University\\
  \texttt{charles\_lovering@brown.edu} \\
  \And
   Ellie Pavlick \\
  Department of Computer Science\\
  Brown University\\
  \texttt{ellie\_pavlick@brown.edu} \\
}

\begin{document}
\definecolor{bg}{rgb}{0.95,0.95,0.95}

\maketitle

\begin{abstract}
Many complex problems are naturally understood in terms of symbolic concepts. For example, our concept of \textit{``cat''} is related to our concepts of \textit{``ears''} and \textit{``whiskers''} in a non-arbitrary way. Fodor (1998) proposes one theory of concepts, which emphasizes symbolic representations related via constituency structures. Whether neural networks are consistent with such a theory is open for debate. We propose unit tests for evaluating whether a system's behavior is consistent with several key aspects of Fodor's criteria. Using a simple visual concept learning task, we evaluate several modern neural architectures against this specification. We find that models succeed on tests of groundedness, modularity, and reusability of concepts, but that important questions about causality remain open. Resolving these will require new methods for analyzing models' internal states.
\end{abstract}

\section{Introduction}

Understanding language requires having representations of the world to which language refers. Prevailing theories in linguistics and cognitive science hold that these representations, or \textit{concepts}, are structured in a compositional way--e.g., the concept of \textit{``car''} can be combined with other concepts (\textit{``gray''}, \textit{``new''})--and that the meanings of composite concepts (\textit{``gray car''}) are inherited predictably from the meanings of the parts. 
State-of-the-art models for natural language processing (NLP) use neural networks (NNs), in which internal representations are points in high-dimensional space. Whether such representations can in principle reflect the abstract symbolic structure presupposed by theories of human language and cognition is an open debate. This paper maintains that the question of whether a model contains the desired type of symbolic conceptual representations is best answered at the \textit{computation level} \citep{marr1982vision}: that is, the diagnostics of ``symbolic concepts'' concern what a system does and why, rather than the details of how that behavior is achieved (e.g., whether it stores vectors vs.\ explicit symbols ``on disk''). Even \citet{fodor1988connectionism}, in their vocal criticism of NNs, assert that ``a connectionist neural network can perfectly well \textit{implement} a classical architecture at the cognitive level''\footnote{Italics added. Here, classical architecture = symbolic architecture, and cognitive level = computational level.}, but do not say how to know if such an implementation has been realized. 

To this end, we propose an API-level specification based on criteria of ``what concepts have to be''  \citep{fodor1998concepts}. Our specification (\S\ref{sec:api}) defines the required behaviors and operations, but is agnostic about implementation. We then consider fully-connectionist systems 
equipped with modern evaluation methods (e.g., counterfactual perturbations, probing classifiers) as candidate systems.
    We present evidence that the evaluated models learn conceptual representations that meet a number of the key criteria (\cref{sec:grounded}--\cref{sec:modular}) but fail on those related to causality (\cref{sec:causal}--\cref{sec:layers}). We argue that more powerful tools for analyzing NNs' internal states may be sufficient to close this gap (\cref{sec:discussion}). 
 Overall, our primary contribution is a framework for seeking converging evidence from multiple evaluation techniques in order to determine whether modern neural models are consistent with a specific theory of concepts. Our experiments offer an updated perspective in the debate about whether neural networks can serve as the substrate of a linguistically competent system. 

\section{``What Concepts Have To Be''}
\label{sec:fodor}

\subsection{Criteria} There is no single agreed-upon standard for what ``concepts'' are \citep{margolis1999concepts}. We base our criteria on those put forth in \citet{fodor1998concepts} as part of a theory which advocates symbolic representations and prioritizes explaining phenomena such as syntactic productivity and semantic compositionality. 
\citet{fodor1998concepts} argues for five conditions required for a conceptual representation to be viable as a model of human-level cognition: \textbf{C1:} ``Concepts are mental\footnote{``Mental'' here implies that the representations are divorceable from the external world. One can token a concept in the absence of relevant perceptual stimuli. E.g., thinking ``If it were raining...'' entails thinking about ``raining'' precisely when it is \textit{not} raining. This distinction is subtle but important. Our unit tests operationalize this via the fact, in 3 out of 4 tests, the perceptual input is held fixed and the intervention is applied to internal state. This is only a first step. Future work will need to explore this issue in more detail to determine what type of perceptual-conceptual distinction suffices to meet this criterion, and how it can be demonstrated empirically.} particulars; specifically, they...function as mental causes and effects''; \textbf{C2:} ``Concepts...apply to things in the world; things in the world ``fall under them''; \textbf{C3:} ``Concepts are constituents of thoughts and...of one another. Mental representations inherit their contents from the contents of their constituents''; \textbf{C4:} ``Quite a lot of concepts [are] learned''; \textbf{C5:} ``Concepts are public...to say that two people share a concept [means] they have tokens of literally the same concept type.''

\subsection{Assumptions and Limitations} 
\label{sec:fodor:limitations}
We focus on \citet{fodor1998concepts}'s criteria since they are concordant with ideas from formal linguistics which have recently been highlighted as weaknesses of NNs \citep{pavlick2022semantic}. 
We don't claim that Fodor's theories should necessarily serve as the standard for NLP systems (indeed, his theories face criticisms). The subset of Fodor's criteria on which we focus (\S\ref{sec:api}) are fairly uncontroversial, and arguably would transfer to alternative theories of conceptual structure--for example, Bayesian causal models \citep{sloman2005causal}. 
We view our tests as necessary but alone insufficient to meet Fodor's criteria. For example, our composite concepts depend on simple conjunction and thus do not address issues about constituency structure in which the argument order matters. Even so, our results offer a valuable starting point on which subsequent theoretical and empirical work can build.

\section{System Specification}\label{sec:api}
We translate key ideas from Fodor's conditions into concrete unit tests for evaluating computational models. Our mapping is not one-to-one: We combine C2 and C5 into a single test focused on whether a concept grounds consistently to perception; we split C3 into two tests and leave aspects to future work; we omit C4 since there is likely little controversy that modern NLP systems ``learn'' concepts. 
Our tests apply to a system holistically, including implementations of diagnostic functions, not just the internal representations. Thus, it is possible for one system to fail our tests, but for a different system with the same internal representation but different implementations of the functions to succeed. See discussions in \cref{sec:limitations} and \cref{sec:discussion}. 

\subsection{Data Types and Basic Functions}
Our domain consists of things in the perceptual world (\texttt{type X}) to which humans assign discrete words (\texttt{type Y}). We follow \citet{fodor1998concepts} in treating word meaning and concepts as interchangeable.\footnote{This is a common assumption. 
Of course, in reality, there are things for which humans may have a concept but do not have the ability to express precisely in language.} Internal concepts may be either \textit{atomic} (without an internal structure) or \textit{composite}, which, in our setting, means they obey a simple conjunctive syntax over atomic constituents (e.g., \textit{``ice''}$\models$\textit{``water''}\&\textit{``solid''}).
We assume two ground-truth functions: \texttt{gt\_label} which returns the name for a given thing and \texttt{gt\_describe} which describes a composite concept (type \texttt{Y}) in terms of its constituents (type \texttt{Set[Y]}).

\begin{figure}[ht!]
\centering
\includegraphics[width=\columnwidth]{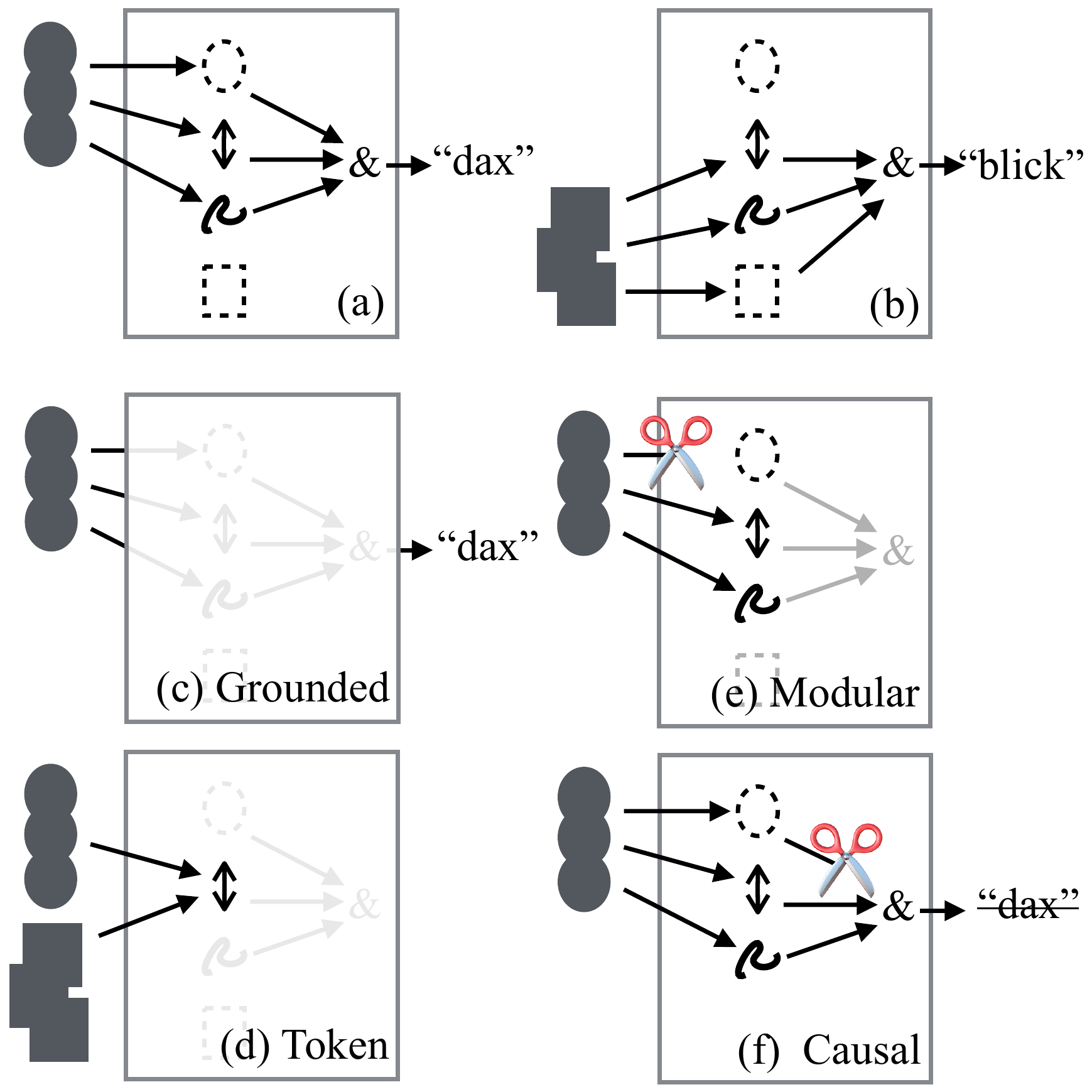}
\caption{Visualization of unit tests as operations on a symbolic graphical model, (a) and (b). (c): Changes in input features lead to expected changes in output. (d): Internal nodes are reused across tokens of the same type. (e): Removing one internal concept does not damage others. (f): Removing internal concepts impacts the model's predictions.}
\label{fig:diagram}
\end{figure}

We require that the system supports an \texttt{encode} operation to map \texttt{X} to an internal representation of type \texttt{Z}, as well as a
\texttt{predict} operation to map \texttt{Z} to \texttt{Y}.
We also require that the system implements two diagnostic functions, i.e., functions unnecessary for the system's usual operations (here, assigning words to inputs), but necessary for measuring properties of the system's internal structure. \texttt{has\_concept} returns true if the system considers the internal representation (\texttt{Z}) to encode a concept (\texttt{Y}); \texttt{ablate} removes the part of the internal representation considered to encode the concept.

\subsection{Unit Tests}

Our specification requires not only that a system supports the above operations, but that its implementation obeys certain constraints, which we formalize via unit tests. Intuitively, it is helpful to think of these unit tests by picturing a symbolic system (e.g., a graphical model) which would pass the tests by construction (Figure \ref{fig:diagram}). In practice, we run our tests on NNs, not graphical models. However, if the models pass our tests, the implication is that the NN has implemented something that, for our purposes, is functionally equivalent to the symbolic model shown in Figure \ref{fig:diagram}.

\subsubsection{\texttt{is\_grounded}}
Our test \texttt{is\_grounded} (analyzed in \cref{sec:grounded}) is derived from the requirements that internal concepts are tied to the external world (C2) in a way that is shared (C5)\footnote{\texttt{is\_grounded} tests whether changes in the input lead to changes in the model's behavior. This is different from Fodor's criteria, which require that the \textit{concept}--i.e., the internal representation--is grounded, and that the representation (not necessarily the behavior) changes in response to external features. We can make this shift because (1) our models' behavior is by definition a function of its internal representation and (2) our test, \texttt{is\_causal}, requires that changes in behavior are explained by changes in the internal representation. Thus success on both \texttt{is\_causal} and \texttt{is\_grounded} entails Fodor's criteria that those things which serve as mental causes and effects are grounded. However, it is plausible that other models could pass using a ``loophole'' in which the behavior is grounded but the internal concepts are not, or could fail due to a technicality in which the representation changes but the model ``decides'' not to change it behavior (though the latter assumes a highly competent system, see \citet{block1981psychologism}).} . Our test requires that models respond to changes in perceptual inputs in the same way that an (idealized) human would respond to those changes, i.e., that \texttt{predict(encode(x)) == gt\_label(x)}. Effectively, this test simply requires a model performs well on the labeling task, but does not care about the representations involved in producing those labels.

\subsubsection{\texttt{is\_token\_of\_type}}

C3 requires that concepts have constituency structure. 
We define two tests which probe aspects of this requirement (see \S\ref{sec:fodor:limitations} for caveats).

 First, \texttt{is\_token\_of\_type} (evaluated in \cref{sec:systematicity}) tests whether different token instances of a concept evaluate to the same semantic type. \citet{fodor1988connectionism} claim this property is required for systematicity and compositionality, arguing that the inference \textit{``Turtles are slower than rabbits$_1$''; ``Rabbits$_2$ are slower than Ferraris''}$\to$\textit{``Turtles are slower than Ferraris''} only follows if, among other things, \textit{``rabbits$_1$''} is treated as the same as (not merely ``similar to'') \textit{``rabbits$_2$''}. We thus require that there exists a computational procedure for mapping models' internal representations into a discrete space, and that this procedure applies in the same way to all token instances of a concept. Concretely, $\forall \texttt{c} \in \texttt{gt\_describe(gt\_label(x))}$ we require that $\texttt{has\_concept(encode(x), c)}$.

\subsubsection{\texttt{is\_modular}}
  Second, \texttt{is\_modular} (evaluated in \cref{sec:modular}) is based on requirements for productivity; e.g., for an NP (e.g., \textit{``John''}) to fit into arbitrarily many contexts (\textit{``John loves Mary''}, \textit{``Joe loves John''}), the representation of the NP must be fully disentanglable from the other words and syntax. We frame this requirement as a test of whether representations support ``slot filling''. That is, given a representation of a composite concept, removal of one constituent concept should produce an unfilled ``slot'' but otherwise leave the remaining constituent concepts intact, i.e., \textit{``\underline{\hspace{3mm}} loves Mary''}.
  Concretely, given \texttt{z = ablate(encode(x), y)}, we require that \texttt{has\_concept(z, y)} is false, and that $\forall \texttt{c} \in \texttt{gt\_describe(gt\_label(x))}$ s.t. \texttt{c $\neq$ y}, \texttt{has\_concept(z, c)} is true.
  
\subsubsection{\texttt{is\_causal}}
Finally, \texttt{is\_causal} (evaluated in \cref{sec:causal}) checks that C1 is met by testing that internal conceptual representations themselves serve as ``mental causes and effects''. As in \citet{fodor1988connectionism}, ``state transitions in Classical machines are causally determined by \textit{the structure--including the constituent structure--of the symbol arrays that the machines transform}: change the symbols and the system behaves quite differently''. To operationalize this, we consider the case in which a model's behavior (e.g., its use of a label) is assumed to be in response to having tokened a composite concept `A\&B'. We require that changes in the representation, such that the constituent concept `A' is no longer tokened (or, that the constituent concept which is tokened is no longer labeled as type `A'), result in corresponding changes in the model's behavior.
In practice, this amounts to requiring that ablating a constituent concept results in expected degradation in model performance. That is,  \texttt{predict(ablate(encode(x), c))} should perform at chance if \texttt{c} is a constituent of \texttt{gt\_label(x)}, and should perform equivalently to \texttt{predict(encode(x))} otherwise.

\section{Implementation}\label{sec:implementation}

Our code, data, and results are available at:\\ \url{bit.ly/unit-concepts-drive}.

\subsection{Functions}
\label{sec:functions}
We implement \textbf{\texttt{encode}} with five different models: three pretrained and two from-scratch\footnote{We do claim pretraining is analogous to human learning. Success on our tests is interesting because it provides an existence proof of \textit{one particular} recipe by which the desired representations arise. This is similar to work on syntax in LMs \citep{linzen2021syntactic}, which is valuable despite LM training being very different from how children learn.}. For pretrained models, we use a residual network trained over ImageNet (\Mc) \citep{he2016deep} and two architectures from CLIP  \citep{radford2021learning}--a vision transformer (\Me) \citep{ dosovitskiy2020image} and a residual network (\Md). For from-scratch models, we use a randomly initialized residual network model (\Mb) and a CNN model\footnote{Four layers: filters=(64, 32, 16, 8), kernels=3, stride=2; batch norm \citep{ioffe2015batch} and ReLU activations.} (\Ma). We use the pretrained encoders with no additional training. For the other models, we finetune on a classification task on our data.

To implement \textbf{\texttt{predict}}, we train linear ``probing classifiers'' \citep{sinha2021masked} over the outputs of \texttt{encode} using 
the Adam optimizer \citep{kingma2014method}. \textbf{\texttt{has\_concept}} is also implemented with linear classifiers. Thus, our system considers the output of \texttt{encode} to ``have'' a concept if a probing model can learn to discriminate instances according to the concept. 

To implement \textbf{\texttt{ablate}}, we use Iterative Nullspace Linear Projection (INLP) \citep{ravfogel2020null}, which repeatedly collapses directions that linearly separate the instances of one concept from those of another. INLP has been used to remove concepts like parts of speech from word representations \citep{elazar2020bert}. 

\begin{figure*}[ht!]
    \centering
    \includegraphics[width=\textwidth]{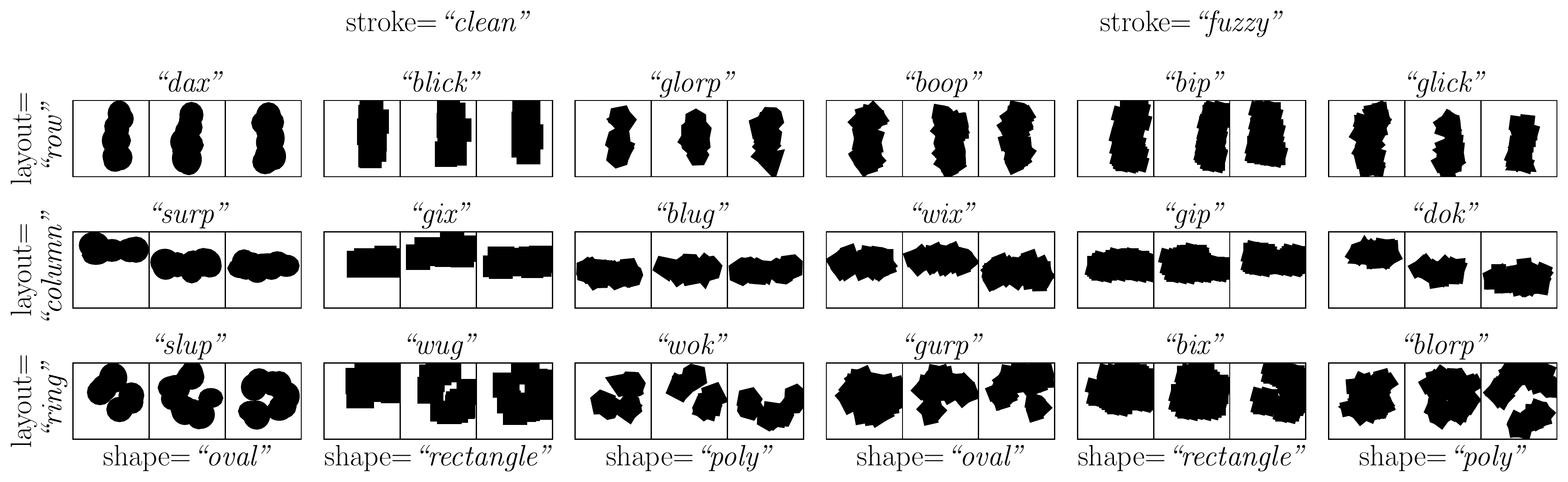}
\caption{\textbf{Dataset.} Three samples from each class; the right nine classes have fuzzy borders (although this admittedly hard to see in these small images.)}
         \label{fig:dataset}
     \end{figure*}

\subsubsection{Limitations}\label{sec:limitations}
We make a few important simplifying assumptions in our implementations, which are necessary in order to employ the available analysis tools at the time of writing. First, since INLP--our implementation for \texttt{ablate}--only removes linear information, we restrict our implementations of \texttt{predict} and \texttt{has\_concept} to be linear models. However, since writing, new methods have been introduced which could in principle be used in place of INLP in our experiments, and would likely yield different results. We discuss possible implications in \cref{sec:causal:discussion}.

Second, in most experiments, we treat the \texttt{encode} function as a block, only analyzing its outputs, rather than ablating concepts in its internal layers. However, looking at individual layers could tell a different story. We provide initial results in \cref{sec:layers}, but a complete investigation warrants significant experiments and is left for future work.

Finally, INLP is iterative, each step removing a direction from the input representation. Our experiments report the results after the first iteration of INLP, as it removes the most salient direction of the concept. Again, future work may find insights in analyzing the removal of subsequent directions.

\subsection{Dataset}\label{sec:dataset}

\subsubsection{Description}

Our \textbf{default dataset} is a synthetic image\footnote{Each image is saved as a PNG and resized to the highest resolution supported by the given model; ImgNet uses 256 x 256 pixels. There is exactly 1 duplicate image in the \textit{colors} dataset for seed = 10.} dataset with 1000 training examples of each of 18 classes, where each class is composed of three from a set of eight \textbf{atomic concepts}  \{3 layouts: horizontal, vertical, ring\} x \{3 shapes: rectangle, oval, polygon\} x \{2 strokes: clean, fuzzy\}. Thus, each class is a \textbf{composite concept} made up of three constituent atomic concepts. See \cref{fig:dataset} for examples.\footnote{
This paper asks whether NNs are consistent with (key aspects of) Fodor’s theory of concepts, \textit{not} whether NNs are equivalent to humans. Synthetic data allows us to study models in a setting where we can guarantee that the desired structure is “correct”. That is, we give Fodor the benefit of the doubt and assume his theory of concepts is correct. Ultimately, we care about the latter question: are NN's human-like? However, in our view, we don’t have the data and theories (just yet) to tackle this in a deep, meaningful way. While Fodor’s theory is certainly \textit{not} a perfect theory of human concepts, at least some aspects of his theory are likely to be present in whatever the “right” theory is, even if not exactly as Fodor envisioned it (e.g., most credible theories appeal to compositionality and causality). Future work can and should relax our generous assumptions, work on non-synthetic data, and analyze NNs through the lens of competing theories of concepts. 
} 

We also create a \textbf{colors dataset} in which the color of the shapes is correlated with the class label. We do this because, in \texttt{is\_grounded} (\cref{sec:grounded}), we find very strong results in the default setting and want to better understand the conditions under which those results hold. The colors dataset emulates a situation where there are spurious features, making it more difficult for a model to ground to the correct perceptual inputs. This dataset is not directly tied to any of Fodor's criteria, but allows us get a more nuanced understanding of our \texttt{is\_grounded} results. Here, each of the 18 classes is correlated with a different color, such that for $p \in \{\textrm{RAND}, 90, 99, 100\}$, a given instance has probability $p$ of expressing that paired color, with remaining $1 - p$ probability distributed uniformly over the other colors. $\textrm{RAND} = 5.6\%$, i.e. $1/18$. 

\subsubsection{Seen and Unseen Examples}\label{sec:seen}
To test the generality of a model's representations, we train the diagnostic functions \texttt{has\_concept} and \texttt{ablate} on a subset of the full 18 classes. We define \textbf{slice} to mean a set of composite concepts that share the same atomic concepts except along a given dimension. For instance, \textit{``dax''}, \textit{``surp''}, \textit{``slup''} form a slice that delineates layout--i.e., the classes differ in layout but otherwise are the same in terms of shape and stroke. All classes that the diagnostic functions are trained on are considered \textbf{seen} and the other classes are considered \textbf{unseen}. We experiment with two training settings, which, like the colors dataset, are not directly tied to Fodor's criteria, but which allow us to tell a more nuanced story about what it takes for models to pass our tests. In the first setting (\textbf{1 slice}), the probes used to implement \texttt{has\_concept} are trained on a dataset with one class per concept. So, in this setting, instances that fall under the concept \textit{``horizontal''} would all be drawn from \textit{``dax''}. In the second setting (\textbf{N-1 slices}), probes are trained on many classes per concept. Here, instances of \textit{``horizontal''} would be drawn from several classes (\textit{``blick''}, \textit{``glorp''}, etc). In \cref{sec:grounded}--\cref{sec:causal}, we focus on the results over unseen classes; performance over seen classes is generally high across all evaluations.

\subsubsection{Human Performance}
We run a Mechanical Turk study with 150 individuals. Subjects are given three exemplars of each class (equivalent to Figure \ref{fig:dataset}), and are then asked to assign a novel instance to one of the 18 classes. Across 1500 predictions, the majority label agrees with our ground truth label 63\% of the time (over a 5.6\% random baseline). We find that mistakes are systematic and predictable: e.g., subjects routinely confusing \textit{``gix''} and \textit{``gip''} as the \textit{``clean''} versus \textit{``fuzzy''} edge is difficult to discern in this setting.

Thus, some of our class distinctions rely on perceptual features that are difficult for humans to distinguish, but which models are able to differentiate well. This is an important discussion point, but does not undermine the validity of the present study. In general, conceptual representation is considered to be divorceable from perception: the fact that one might mistake a cat for a skunk does not mean they do not have the concept of \texttt{cat}. By similar logic, the fact that our models have super-human perception in this domain need not prevent us from analyzing the structure of the concepts that they represent, or comparing them to a ground truth that imagines humans to have perfect perception.

\section{Test 1: Predictions are Grounded}\label{sec:grounded}
\texttt{Is\_grounded} requires that if, definitionally, the difference between \textit{``dax''} and \textit{``blick''} is roundness, then this visual attribute should dictate predictions. 

\subsection{Experimental Design}
We use counterfactual minimal pairs, which have  been used in both NLP \citep{huang-etal-2020-reducing} and computer vision \citep{goyal2019counterfactual}. Our dataset (\cref{sec:dataset}) is generated using a set of background parameters (i.e., locations and sizes of the underlying shapes) in addition to the atomic concepts (shape, stroke, and layout). To generate minimal pairs, we sample 1000 sets of these background parameters, and then render each sampled set of parameters for every combination of shape$\times$stroke$\times$layout. This ensures the instances in a pair are equivalent in all visual features (e.g., total surface area covered by shapes, relative distance between shapes, etc) except those features which change as a direct consequence of manipulating the target atomic concept. We generate minimal pairs in the colors dataset (\cref{sec:dataset}) in the same way, treating color as another background parameter. 
After setting up the minimal pairs, we measure the probability that \texttt{predict(encode(}$\cdot$\texttt{))} == \texttt{gt\_label(}$\cdot$\texttt{)}.

If the model grounds concepts to the desired perceptual features, then it should perform perfectly at classifying the images across all settings. If the model performs poorly, we interpret this as evidence that the model grounds the concept to some features in a way that would not be ``shared'' with (idealized) humans, e.g., the model considers \textit{``dax''} to ground to color or size of shapes, rather than solely to \textit{``circle''}\&\textit{``horizontal''}\&\textit{``smooth''}. 

\subsection{Results}
\begin{figure}[!ht]
    \centering
    \begin{tikzpicture}
        \begin{axis}[
            xtick={1,2,3,4,5},
            xticklabels={default, $\textrm{RAND}$, $90\%$, $99\%$, $100\%$},
            legend columns=3,
            legend style={at={(1,-0.3)},anchor=north east,font=\footnotesize},
            width=\columnwidth,
            height=1.5in,
            ylabel=Accuracy
        ]
  
\addplot+[color=C1, mark = triangle*, mark options={fill=C1}] coordinates { (1, 0.99) (2, 0.99) (3, 0.95) (4, 0.78) (5, 0.33) };
\addplot+[color=C2, mark = square*, mark options={fill=C2}] coordinates { (1, 1.00) (2, 0.98) (3, 0.94) (4, 0.73) (5, 0.33) };
\addplot+[color=C3, mark = diamond*, mark options={fill=C3}] coordinates { (1, 0.98) (2, 0.97) (3, 0.94) (4, 0.76) (5, 0.45) };
\addplot+[color=C4, mark = pentagon*, mark options={fill=C4}] coordinates { (1, 0.98) (2, 0.85) (3, 0.32) (4, 0.23) (5, 0.10) };
\addplot+[color=C5, mark = *, mark options={fill=C5}] coordinates { (1, 1.00) (2, 1.00) (3, 0.87) (4, 0.44) (5, 0.10) };
\addplot+[color=C6,dashed, mark=star, mark options={fill=C6}] coordinates { (1, 0.06) (2, 0.06) (3, 0.06) (4, 0.06) (5, 0.06) };

\legend{\Me, \Md, \Mc, \Mb, \Ma, chance};
\end{axis}
\end{tikzpicture}
    \caption{Results for \texttt{is\_grounded} on the colors dataset. Performance for all models degrades when trained on data in which color is spuriously correlated with the target concepts, and then tested on out-of-distribution minimal pairs. However, pretrained models still perform well above chance.}
    \label{fig:t1}
\end{figure}
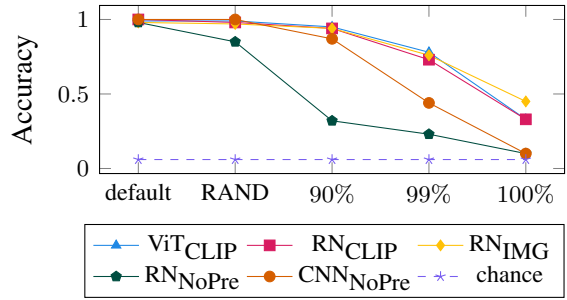

The models perform well on the default dataset ($\sim98\%$). When the classes are highly correlated with a spurious color feature, performance degrades (Figure \ref{fig:t1}).
However, notably, even when models are trained on highly imbalanced data (e.g., with 99\% of \textit{``dax''}s being red), the pre-trained models still perform well above random out-of-distribution (75\% over a 5.6\% random baseline). 

\begin{figure*}[ht!]
         \centering 
\input{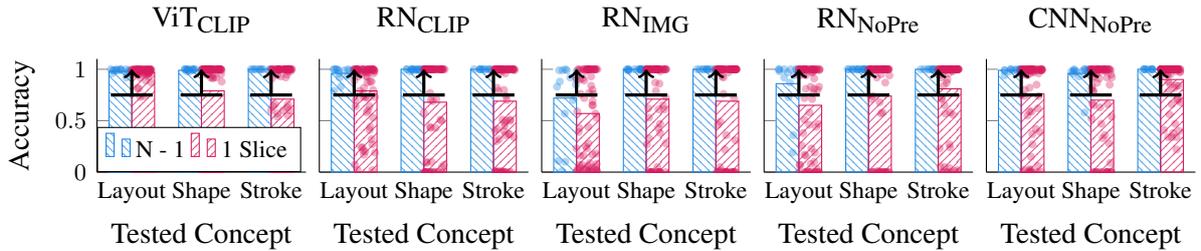}
         \caption{\textbf{\texttt{is\_token\_of\_type} on unseen classes.} The points show the accuracies over seeds and the unseen test classes; the bar shows the mean over these points. Black arrows indicate expectations--we want to see models performing well, as high accuracy is indicative of a reusable type representation that generalizes to unseen concepts.}
         \label{fig:t2}
     \end{figure*}

\subsection{Discussion}\label{sec:grounded:discussion}
We interpret this as a positive result: The results on the default dataset demonstrate that the pretrained models' behavior is explained by the expected perceptual features, satisfying \texttt{is\_grounded}. The degradation in performance when using the colors dataset raises two issues worthy of discussion. First, across our unit tests, this result is the one of only places in which we see a real difference between pretrained and from-scratch models. These results suggest that the pretrained models (which have been trained with access to linguisitic information, i.e., category labels for ImageNet and captions for CLIP) encode an inductive bias for shape over color. That is, even in the setting in which color is perfectly correlated with the class label, the models still generalize based on shape rather than color around half of the time. Such findings echo previously published arguments that pretraining can encode inductive biases that help models learn language more efficiently \citep{lovering2021predicting,warstadt2020learning,mueller2022coloring}.

Second, while poor out-of-distribution generalization is not desireable, it is important to emphasize that it is \textit{not}
inconsistent with the use of symbolic concepts. For example, a model which explicitly represents symbols (e.g., Naive Bayes) could exhibit a similar drop in performance as the prior given the correlation in the training data makes the correct class less likely. As written, Fodor's criteria do not adjudicate on this issue. Thus, with respect to grounding, fully characterizing neural networks in terms of their symbolic representations (or lack thereof) requires refined criteria which can discriminate between models which represent grounded symbols (but make errors in learning) from models that do not represent grounded symbols at all.

\section{Test 2: Representations Encode Types}\label{sec:systematicity}
\texttt{Is\_token\_of\_type} requires that the system's representations of concepts can be mapped to discrete types in a reusable way.

\subsection{Experimental Design}
We train \texttt{has\_concept} on a subset of the slices from the dataset (see \cref{sec:seen}). For example, we can train \texttt{has\_concept} to predict the layout (\textit{``vertical''}, \textit{``horizontal''}, or \textit{``ring''}) by training it on examples of \textit{``dax''}, \textit{``surp''} and \textit{``slup''}, which differ only in the layout constituent, but are identical in the other constituents, (\textit{``oval''}, \textit{``smooth''}). We then evaluate on unseen classes, such as \textit{``blick''}, \textit{``gix''} and \textit{``wug''}, which exemplify the same variation in layout, but do so in the context of other constituents not seen in training (e.g., \textit{``rectangle''}).

We take good generalization as evidence that the model's representations of a concept can be viewed as tokens of the same concept type. For example, whenever the model receives an input that falls under the concept \textit{``vertical''}, the concept of \textit{``vertical''} is tokened in the model's internal representations in a way which can be reliably localized by a single, fixed \textit{``vertical''}-type detector. Generalization to unseen classes indicates that the tokening of \textit{``vertical''} is not dependent on the other concepts that might be tokened simultaneously (e.g., \textit{``oval''} or \textit{``rectangle''}). Poor generalization suggests that models' internal representations are context dependent: \textit{``vertical''} in the context of \textit{``oval''} is not of the same type as \textit{``vertical''} in the context of \textit{``rectangle''}.

\subsection{Results}
The results are overall positive. All models show near-perfect accuracies on seen classes ($>99\%$, not shown). Over the unseen classes (\cref{fig:t2}), the models perform better in the easier {N-1 slices} setting (when generalizing from 15 seen classes to 3 unseen classes). For {1 slice}, the accuracies are lower but still well above chance--around to 75\%.
      
 \begin{figure*}[ht!]
         \centering 
\input{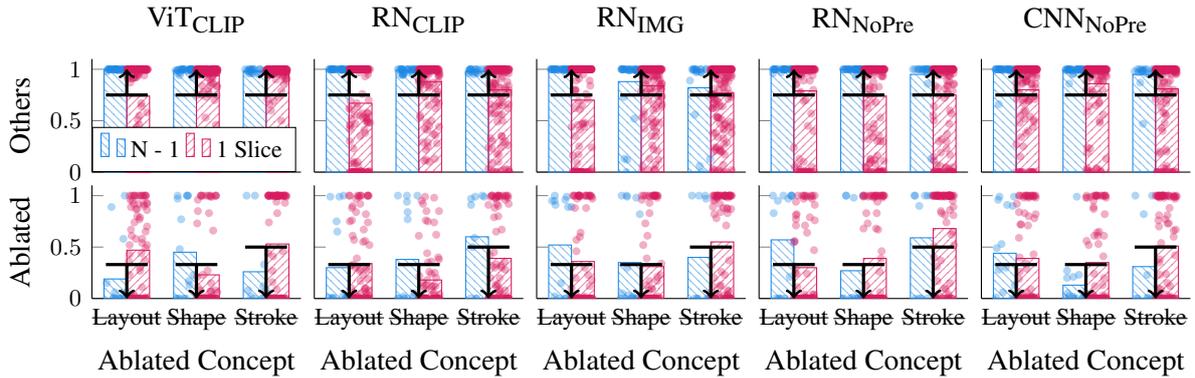}
         \caption{\textbf{\texttt{is\_modular} on unseen classes.} Arrows indicate expectations: performance for the ablated concepts (top row) should be at or below random and performance on other concepts (bottom row) should be high. Points show the accuracies over seeds and  unseen test classes; the bar shows the mean. } 
         \label{fig:t3-chart}
     \end{figure*}

\subsection{Discussion}
\label{sec:systematicity:discussion}
Overall, representations of atomic concepts appear to be ``the same'' across contexts, generalizing well to unseen compositions. The performance differential between {1 slice} and {N - 1 slice} suggests (intuitively) that more varied data enables the \texttt{has\_concept} probe to better identify the stable, defining features of the concept: i.e., seeing \textit{``vertical''} in the context of both \textit{``oval''} and \textit{``rectangle''} makes it easier to recognize \textit{``vertical''} in the context of previously-unseen \textit{``polygon''}. As was the case with the out-of-distribution generalization results discussed in \cref{sec:grounded:discussion}, these results about the amount and variety of training data required are interesting, but do not speak directly to the question of symbolic representations. Rather, our results on 1 slice vs.\ N - 1 slices correspond to a question about acquisition, and is an issue on which Fodor's criteria are silent. Other theories of concepts focus on acquisition \citep{spelke2007core,carey2009origin} and make empirical predictions about the amount and distribution of data from which certain concepts should be acquirable. Future work could expand our unit tests to reflect such empirical predictions, in addition to the in-principle criteria proposed by Fodor. 

\section{Test 3: Representations are Modular}\label{sec:modular}
\texttt{Is\_modular} tests that removing one constituent concept from the representation of a composite concept does not harm the other constituents.\footnote{Whether concepts should be entangled, i.e., ``holism'' \cite{sep-meaning-holism}, is an area of extensive debate. We make some strong assumptions following Fodor's ideals. See Footnote 8.}

\subsection{Experimental Design}
We use \texttt{ablate} to remove a given constituent and then assert that \texttt{has\_concept} is unable to detect the removed concept, but still able to detect the remaining constituents. For example, \textit{``dax''} $\models$ \textit{``oval''\&``horizontal''\&``smooth''} is a composite concept. We require that ablating \textit{``horizontal''} from a tokened representation of \textit{``dax''} results in a representation of the form \textit{``oval''}\&\underline{\hspace{3mm}}\&\textit{``smooth''}, which leaves the layout ``slot'' empty, but otherwise preserves the information about the structure and type of the composition. In our implementation, without loss of generality, we ablate sets of atomic concepts (e.g., ablating all three layout concepts together) rather than a single concept at a time. 

High accuracy on the ablated concept means the system failed to implement \texttt{ablate} correctly. Low accuracy on the concepts that were not ablated (e.g., if removing layout means \texttt{has\_concept} no longer can distinguish \textit{``rectangle''} from \textit{``oval''}) means that constituent representations are entangled in a way likely incompatible with e.g., productivity.
Thus, for each atomic concept dimension (layout, shape, stroke) we run three tests--one to check that performance at detecting the ablated concepts is low and two to check that performance at detecting the other two dimensions is high. We consider ``high'' to be >75\% accuracy\footnote{This threshold is arbitrary, but allows us to talk in terms of explicit pass/fail criteria for our unit tests.}; random is 33\% for layout and shape, and 50\% for stroke.

\begin{figure*}[ht!]
\centering
\input{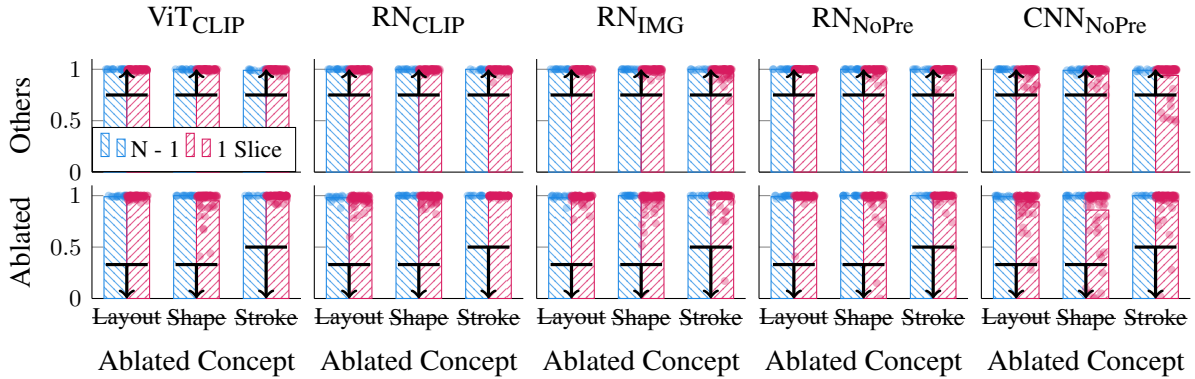}
\caption{\textbf{\texttt{is\_causal} on unseen classes.}  Arrows indicate expectations: performance for the ablated concepts (top) should be at chance and performance on other concepts (bottom) should be high. Bars show mean accuracies over seeds and  unseen test classes. Accuracies are over classes (composite concepts). }  
\label{fig:t4-chart}
\end{figure*}

\subsection{Results}
All the models are largely successful (\cref{fig:t3-chart}). Overall, performance is low on the removed concept but high on the remaining concepts, as desired. Performance is higher variance in the harder 1 slice setting. For example, when layout is ablated in \Me, the accuracy for detecting layout is far below random in the N - 1 slice setting, but marginally above random in the 1 slice setting.
\subsection{Discussion}
\label{sec:modular:discussion}

Across models and training configurations, the trends are in the expected direction: performance on the ablated concept is low (near random) and performance on other concepts is high. In the harder 1-slice setting, performance on the not-ablated concepts sometimes degrades, meaning, for example, its not possible to remove the constituent \textit{``vertical''} from \textit{``dax''} without also damaging the representation of \textit{``oval''} to some extent. In terms of Fodor's criteria for constituency, this suggests a problem, as the lack of modularity would make it difficult to explain phenomena such as infinite productivity--i.e., if \textit{``oval''} cannot be fully divorced from \textit{``vertical''}, it becomes difficult to explain how the same \textit{``oval''} is able to combine with arbitrarily many different layouts (\textit{``horizontal''}, \textit{``ring''}, etc). However, the evidence is hardly damning--the patterns are largely consistent with expectations. As in \cref{sec:systematicity:discussion}, this represents a direction in need of future work and discussion. These results could become unambiguously positive if we concede that models might require sufficient training in order to learn modular concept representations. Fodor's theory does not offer criteria for what is ``sufficient'', but subsequent experiments could draw on other theories from developmental psychology to determine such criteria, and then refine the unit tests accordingly.

\section{Test 4: Representations are \textit{Not} Causal}\label{sec:causal}
\texttt{Is\_causal} tests that the internal representations serve as ``mental causes and effects''. 
Where \texttt{is\_token\_of\_type} and \texttt{is\_modular} demonstrated that models' representations can be labeled and manipulated according to discrete types, we now test that those types are causally implicated in model behavior--e.g., if the constituent concept \textit{``oval''} is no longer tokened, will this prevent the model from producing the label \textit{``dax''}? Similar to \texttt{is\_grounded}, this test relies on counterfactual perturbations, but differs in that the perturbations are applied to the model's internal representations, rather than to the perceptual input.

\subsection{Experimental Design}
We evaluate \texttt{predict} after removing a concept with \texttt{ablate}. We expect this to impair the model's ability to reason about the ablated concept, but not others. For example, if we remove the layout dimension, the model should be able to distinguish between \textit{``blick''} and \textit{``dax''} (as they differ in shape), but unable to distinguish between \textit{``blick''} and \textit{``slup''} (as they differ in layout). We thus distinguish two measures of accuracy: the rate at which the model's predicted concept matches the true concept along the removed dimension (which should be at random), and the rate at which the model's predicted concept matches the true concept along the other dimensions (which should be high). We take >75\% accuracy to be high; random is 33\% for layout and shape, and 50\% for stroke.

\subsection{Results}
All of our models fail this test (\cref{fig:t4-chart}). Accuracies with respect to the ablated features stay far above random. 
The pattern holds whether we train on 1 or N-1 slices, and whether we evaluate on seen (not shown) or unseen classes. Increasing the iterations of INLP (\cref{sec:implementation}) (not shown) causes performance to deteriorate for all concepts (even those which we are not trying to ablate), a different pattern which nonetheless constitutes a failure on our unit test.

\subsection{Discussion}
\label{sec:causal:discussion}

\begin{figure*}[ht!]
         \centering 
\begin{tikzpicture}
\begin{groupplot}[group    style={columns=5},
    view={0}{90},
    width=2.25cm,
    height=2.25cm,
    scale only axis,
axis y line*=left, 
axis x line*=bottom,
    group style={
    group size=5 by 1,
    horizontal sep=5pt,
    vertical sep=5pt,
    x descriptions at=edge bottom,
    y descriptions at=edge left},
    ymin=0, ymax=1.1,
legend,
    legend style={at={(1.1,0)},anchor=south west,font=\footnotesize},
        legend cell align={left},
    ]        

        \nextgroupplot[
            ylabel=Accuracy,
            title=\Me,
            xticklabel style = {rotate=90},
        xmin=.5,xmax=5.5,
        xtick={ 1,2,3,4,5 },
                    xticklabels={ conv,layer 1,layer 2,layer 3,layer + }]
            \addplot+[color=C1, mark = triangle*, mark options={fill=C1}] coordinates { (1, 0.98)(2, 0.99)(3, 0.99)(4, 1.00)(5, 1.00) };
            \addplot+[color=C2, mark = square*, mark options={fill=C2}] coordinates { (1, 1.00)(2, 1.00)(3, 1.00)(4, 1.00)(5, 1.00) };
            \addplot+[color=C3, mark = diamond*, mark options={fill=C3}] coordinates { (1, 0.63)(2, 0.94)(3, 0.99)(4, 1.00)(5, 1.00) };
            \addplot+[color=C5, mark = *, mark options={fill=C5}] coordinates { (1, 0.48)(2, 0.81)(3, 0.93)(4, 0.99)(5, 0.99) };
            \addplot+[color=C6,dashed, mark=star, mark options={fill=C6}] coordinates { (1, 0.42)(2, 0.73)(3, 0.83)(4, 0.96)(5, 0.99) };
            
        \nextgroupplot[
            title=\Md,
            xticklabel style = {rotate=90},
        xmin=.5,xmax=8.5,
        xtick={ 1,2,3,4,5,6,7,8 },
                    xticklabels={ conv 1,conv 2,conv 3,layer 1,layer 2,layer 3,layer 4,attnpool }]
            \addplot+[color=C1, mark = triangle*, mark options={fill=C1}] coordinates { (1, 0.78)(2, 0.93)(3, 0.94)(4, 0.78)(5, 0.93)(6, 0.95)(7, 0.98)(8, 0.99) };
            \addplot+[color=C2, mark = square*, mark options={fill=C2}] coordinates { (1, 1.00)(2, 1.00)(3, 1.00)(4, 1.00)(5, 1.00)(6, 1.00)(7, 1.00)(8, 1.00) };
            \addplot+[color=C3, mark = diamond*, mark options={fill=C3}] coordinates { (1, 0.50)(2, 0.66)(3, 0.63)(4, 0.72)(5, 0.56)(6, 0.63)(7, 0.67)(8, 1.00) };
            \addplot+[color=C5, mark = *, mark options={fill=C5}] coordinates { (1, 0.18)(2, 0.57)(3, 0.46)(4, 0.23)(5, 0.26)(6, 0.35)(7, 0.61)(8, 1.00) };
            \addplot+[color=C6,dashed, mark=star, mark options={fill=C6}] coordinates { (1, 0.19)(2, 0.41)(3, 0.34)(4, 0.33)(5, 0.26)(6, 0.30)(7, 0.44)(8, 0.99) };
            
        \nextgroupplot[
            title=\Mc,
            xticklabel style = {rotate=90},
        xmin=.5,xmax=5.5,
        xtick={ 1,2,3,4,5 },
                    xticklabels={ conv,layer 1,layer 2,layer 3,avgpool }]
            \addplot+[color=C1, mark = triangle*, mark options={fill=C1}] coordinates { (1, 0.99)(2, 0.97)(3, 0.99)(4, 0.98)(5, 0.95) };
            \addplot+[color=C2, mark = square*, mark options={fill=C2}] coordinates { (1, 1.00)(2, 1.00)(3, 1.00)(4, 1.00)(5, 1.00) };
            \addplot+[color=C3, mark = diamond*, mark options={fill=C3}] coordinates { (1, 0.68)(2, 0.66)(3, 0.63)(4, 0.75)(5, 1.00) };
            \addplot+[color=C5, mark = *, mark options={fill=C5}] coordinates { (1, 0.66)(2, 0.42)(3, 0.54)(4, 0.65)(5, 1.00) };
            \addplot+[color=C6,dashed, mark=star, mark options={fill=C6}] coordinates { (1, 0.45)(2, 0.37)(3, 0.43)(4, 0.49)(5, 0.95) };
            
        \nextgroupplot[
            title=\Mb,
            xticklabel style = {rotate=90},
        xmin=.5,xmax=5.5,
        xtick={ 1,2,3,4,5 },
                    xticklabels={ conv,layer 1,layer 2,layer 3,avgpool }]
            \addplot+[color=C1, mark = triangle*, mark options={fill=C1}] coordinates { (1, 0.99)(2, 0.97)(3, 0.96)(4, 0.90)(5, 0.97) };
            \addplot+[color=C2, mark = square*, mark options={fill=C2}] coordinates { (1, 1.00)(2, 1.00)(3, 1.00)(4, 1.00)(5, 1.00) };
            \addplot+[color=C3, mark = diamond*, mark options={fill=C3}] coordinates { (1, 0.66)(2, 0.81)(3, 0.89)(4, 0.98)(5, 1.00) };
            \addplot+[color=C5, mark = *, mark options={fill=C5}] coordinates { (1, 0.55)(2, 0.71)(3, 0.89)(4, 0.91)(5, 1.00) };
            \addplot+[color=C6,dashed, mark=star, mark options={fill=C6}] coordinates { (1, 0.42)(2, 0.51)(3, 0.57)(4, 0.56)(5, 0.97) };
            
        \nextgroupplot[
            title=\Ma,
            xticklabel style = {rotate=90},
        xmin=.5,xmax=5.5,
        xtick={ 1,2,3,4,5 },
                    xticklabels={ conv,layer 1,layer 2,layer 3,dense }]
            \addplot+[color=C1, mark = triangle*, mark options={fill=C1}] coordinates { (1, 0.98)(2, 0.98)(3, 0.98)(4, 0.99)(5, 1.00) };
            \addplot+[color=C2, mark = square*, mark options={fill=C2}] coordinates { (1, 1.00)(2, 1.00)(3, 1.00)(4, 1.00)(5, 1.00) };
            \addplot+[color=C3, mark = diamond*, mark options={fill=C3}] coordinates { (1, 0.56)(2, 0.55)(3, 0.58)(4, 0.63)(5, 1.00) };
            \addplot+[color=C5, mark = *, mark options={fill=C5}] coordinates { (1, 0.29)(2, 0.48)(3, 0.49)(4, 0.57)(5, 1.00) };
            \addplot+[color=C6,dashed, mark=star, mark options={fill=C6}] coordinates { (1, 0.29)(2, 0.36)(3, 0.36)(4, 0.40)(5, 0.99) };
            
\legend{Layout, Shape, Stroke, Direct, Composed};
\end{groupplot}
\end{tikzpicture}
         \caption{\textbf{Probing performance explains downstream performance across layers.} Composed Probes: accuracy that would result by directly composing the predictions of the probes for each constituent concept; Direct Classification: accuracy of a classifier trained at the given layer to predict the \textit{composite} concept. The remaining lines show the probing performance for the constituent concepts.}
         \label{fig:layers}
     \end{figure*}
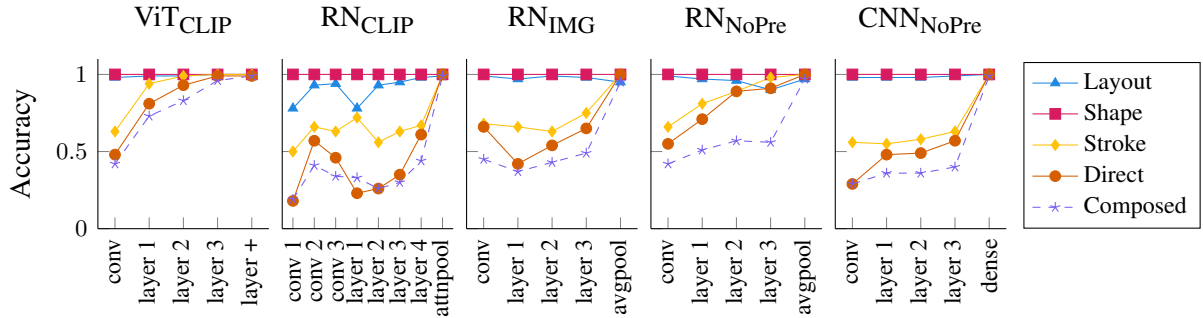

These models in general pass \texttt{is\_modular}, meaning that there exists a localizable representation of each atomic concept.
Thus, this subsequent failure suggests that \texttt{predict} ends up using different representations than those which are used by \texttt{has\_concept}. That is, while there exists a part of the internal representation that encodes the atomic concepts, \texttt{predict} relies on a different part of the internal representation to make decisions about composite concepts.

One possible explanation for this result is that the model tokens \textit{both} the atomic concepts and the composite ones simultaneously, with each concept (composite or not) represented as its own symbol, and \texttt{predict} uses only the composite ones directly. For example, observing an instance of \textit{``dax''} causes the model to token the atomic \textit{``oval''} and \textit{``horizontal''} but also a composite concept \textit{``oval''}\&\textit{``horizontal''} which is a symbol in and of itself. Whether or not such behavior is consistent with Fodor's criteria depends on the causal relationship between these tokenings--i.e., does tokening \textit{``oval''}\&\textit{``horizontal''} entail tokening \textit{``oval''}? Future work could answer this question by looking more closely at the way representations evolve during training or across layers during processing. We present initial investigations on the latter in \cref{sec:layers}.

Finally, as discussed in \cref{sec:api}, our specification applies not just to the representations, but to the system as a whole. 
Thus, the implementation of \texttt{ablate} (INLP in our case), is part of the evaluated system. When a model fails this test, we cannot say whether there was a critical flaw with the representation or rather that the concept ablation itself failed--e.g., because of assumptions of linearity, of treating \texttt{encode} as a block, etc. It is possible that, if new techniques are used to instantiate \texttt{ablate}, the same representations might fare better (or worse) according to our tests. For example, since writing, new techniques for applying non-linear perturbations \citep{tucker2021if,meng2022locating} have been proposed. Such methods could potentially be incorporated into our framework to yield new insights on this particular test.

\begin{figure*}[ht!]
         \centering 
\begin{tikzpicture}
\begin{groupplot}[group    style={columns=5},
    view={0}{90},
    width=2.25cm,
    height=2.25cm,
    scale only axis,
axis y line*=left, 
axis x line*=bottom,
    group style={
    group size=5 by 1,
    horizontal sep=5pt,
    vertical sep=5pt,
    x descriptions at=edge bottom,
    y descriptions at=edge left},
    ymin=0, ymax=1.1,
legend,
    legend style={at={(1.1,0)},anchor=south west,font=\footnotesize},
        legend cell align={left},
    ]        

        \nextgroupplot[
            ylabel=Accuracy,
            title=\Me,
            xticklabel style = {rotate=90},
        xmin=.5,xmax=5.5,
        xtick={ 1,2,3,4,5 },
                    xticklabels={ conv,layer 1,layer 2,layer 3,layer + }]
            \addplot+[color=C4, mark = pentagon*, mark options={fill=C4}] coordinates { (1, 0.61)(2, 0.79)(3, 0.85)(4, 0.94)(5, 0.98) };
            \addplot+[color=C6,dashed, mark=star, mark options={fill=C6}] coordinates { (1, 0.42)(2, 0.73)(3, 0.83)(4, 0.96)(5, 0.99) };
            
        \nextgroupplot[
            title=\Md,
            xticklabel style = {rotate=90},
        xmin=.5,xmax=8.5,
        xtick={ 1,2,3,4,5,6,7,8 },
                    xticklabels={ conv 1,conv 2,conv 3,layer 1,layer 2,layer 3,layer 4,attnpool }]
            \addplot+[color=C4, mark = pentagon*, mark options={fill=C4}] coordinates { (1, 0.34)(2, 0.55)(3, 0.53)(4, 0.37)(5, 0.51)(6, 0.57)(7, 0.55)(8, 0.98) };
            \addplot+[color=C6,dashed, mark=star, mark options={fill=C6}] coordinates { (1, 0.19)(2, 0.41)(3, 0.34)(4, 0.33)(5, 0.26)(6, 0.30)(7, 0.44)(8, 0.99) };
            
        \nextgroupplot[
            title=\Mc,
            xticklabel style = {rotate=90},
        xmin=.5,xmax=5.5,
        xtick={ 1,2,3,4,5 },
                    xticklabels={ conv,layer 1,layer 2,layer 3,avgpool }]
            \addplot+[color=C4, mark = pentagon*, mark options={fill=C4}] coordinates { (1, 0.59)(2, 0.59)(3, 0.60)(4, 0.61)(5, 0.97) };
            \addplot+[color=C6,dashed, mark=star, mark options={fill=C6}] coordinates { (1, 0.45)(2, 0.37)(3, 0.43)(4, 0.49)(5, 0.95) };
            
        \nextgroupplot[
            title=\Mb,
            xticklabel style = {rotate=90},
        xmin=.5,xmax=5.5,
        xtick={ 1,2,3,4,5 },
                    xticklabels={ conv,layer 1,layer 2,layer 3,avgpool }]
            \addplot+[color=C4, mark = pentagon*, mark options={fill=C4}] coordinates { (1, 0.62)(2, 0.62)(3, 0.65)(4, 0.64)(5, 0.98) };
            \addplot+[color=C6,dashed, mark=star, mark options={fill=C6}] coordinates { (1, 0.42)(2, 0.51)(3, 0.57)(4, 0.56)(5, 0.97) };
            
        \nextgroupplot[
            title=\Ma,
            xticklabel style = {rotate=90},
        xmin=.5,xmax=5.5,
        xtick={ 1,2,3,4,5 },
                    xticklabels={ conv,layer 1,layer 2,layer 3,dense }]
            \addplot+[color=C4, mark = pentagon*, mark options={fill=C4}] coordinates { (1, 0.59)(2, 0.61)(3, 0.60)(4, 0.53)(5, 0.98) };
            \addplot+[color=C6,dashed, mark=star, mark options={fill=C6}] coordinates { (1, 0.29)(2, 0.36)(3, 0.36)(4, 0.40)(5, 0.99) };
            
\legend{NMI, Composed};
\end{groupplot}
\end{tikzpicture}
         \caption{\textbf{Expected vs observed mistakes are different.} Mutual information between the probing and downstream predictions at the instance level. If there were a direct causal connection between the constituent concept and the composite prediction, we would expect high NMI across all layers. Instead, for most models, NMI is only high at the final layer.} 
         \label{fig:nmi}
     \end{figure*}
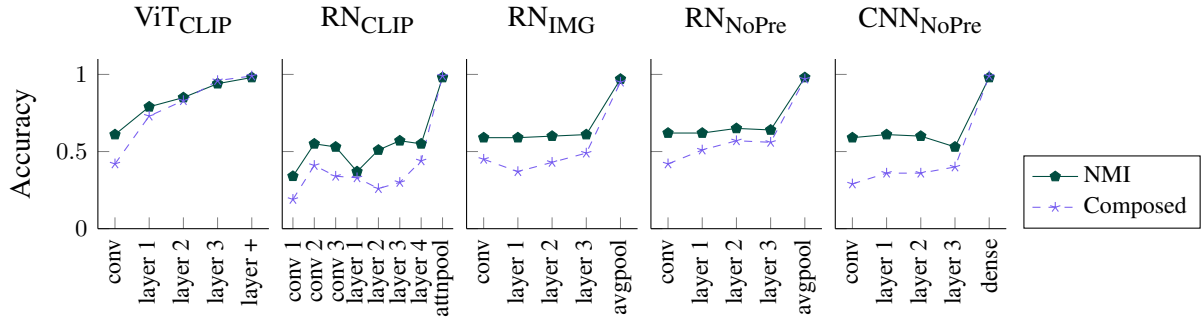

\section{Analysis: Concepts Across Layers}\label{sec:layers}

\subsection{Hypothesis}
Here, we conduct a preliminary investigation into one hypothesis about the reason for our models' failure on \texttt{is\_causal}. Specifically, we hypothesize that the causal structure exists, but it unfolds across layers. The constituent concepts (e.g., \textit{``oval''} and \textit{``horizontal''}) are tokened in early layers, and are subsequently composed such that the composite concept (\textit{``dax''} = \textit{``oval''}\&\textit{``horizontal''}) is tokened at the final layer as its own symbol and is the direct effect of the model's predicted label. Below, we investigate two predictions of this hypothesis, and observe mixed results.

\subsection{Aggregate Analysis}
If our hypothesis is true, we would expect to see 1) that concepts should emerge in the expected order across layers, i.e., constituent concepts before composite concepts and 2) errors in labeling the composite concept at a given layer should be explained by errors in identifying the constituents at that layer. That is, if the model cannot recognize \textit{``oval''} vs.\ \textit{``rectangle''} until layer 4, it should not be able to differentiate \textit{``dax''} from \textit{``blick''} (which depend on the shape distinction) before that layer. Moreover, if the model's failure to recognize \textit{``oval''} vs.\ \textit{``rectangle''} is the reason for the mislabel, the observed error in labeling the composite concepts should be equal to the product of the errors the constituents. That is, considering \textit{``dax''}s, if errors in the constituents cause errors in the composite, the model should mislabel \textit{``dax''} as \textit{``blick''} exactly as often as \texttt{has\_concept} mistakenly returns \textit{``rectangle''} instead of \textit{``oval''}.

\cref{fig:layers} shows predictions from probing models for each concept at each layer. It also shows the \textit{composed probe} accuracy, computed by combining the predictions of each of the probing classifiers, as well as the \textit{direct classification} accuracy, computed by measuring the performance of a new classifier trained to predict the final class \textit{at each layer}.\footnote{Because CNNs and ViTs have multiple dimensions, to get a vector representation for a given layer, we mean across the channels and then flatten into a vector. There are many other possible approaches we did not evaluate.} The trend is promising: composite concepts are recognized only after constituents are recognized, and, for most models, the direct classification accuracy is close to what we expect based on composed probes (though often slightly higher, especially on the from-scratch models).\footnote{n.b. In early layers, models make the same mistakes people do, e.g., confusing fuzzy ovals and polygons (\cref{sec:dataset}).}

\subsection{Instance-Level Analysis}

If our hypothesis holds, not only should the error rates be similar, but the direct class prediction should be  predicted by the composed probes. That is, if at a given layer, the model is given an image of a \textit{``dax''} and mistakenly detects \textit{``rectangle''} (according to the probe) instead of \textit{``oval''}, then the model should label the input as \textit{``blick''}.

To quantify whether the instance-level predictions behave this way, we compute the normalized pointwise mutual information (NMI, which ranges from $0$ to $1$) between the direct prediction and the composition of the probe predictions. If the direct prediction is indeed a function of the constituent probes, we would expect to see high NMI (near $1.0$) across the board--i.e., even when the model's accuracy is low, the NMI would be high if it was erring in the expected way. However, \cref{fig:nmi} shows there is relatively little mutual information until the final layer of the network (\Me might be an exception). In other words, while the probing and downstream models have similar error rate in aggregate, they make \textit{different mistakes} on individual instances.

This result is inconclusive: while high NMI would have been suggestive of a causal connection between the probes and the classifier, low NMI doesn't necessarily mean such a link does not exist. E.g., if a model is altogether failing to differentiate \textit{``rectangle''}s and \textit{``oval''}s, and thus failing to differentiate  \textit{``dax''}s and  \textit{``blicks''}s, then both the probe and the classifier might resort to pure guessing between these labels, and thus appear to disagree even though they in fact depend on the same (underdetermined) conceptual representation.

\section{Summary}\label{sec:discussion}
Overall, our experiments suggest that models exhibit grounded behavior and possess conceptual representations that encode modular, context-independent types. However, we don't find evidence of a direct causal connection between the representations of constituent concepts and those of composite concepts, an essential feature of Fodor's theory on which our specification is based. Our discussions of each idividual experiment (\cref{sec:grounded:discussion},  \cref{sec:systematicity:discussion}, \cref{sec:modular:discussion}, \cref{sec:causal:discussion}) together raise several general themes. 

First, success on our tests often depends on granting assumptions about how concepts are acquired; viz., how should concepts be learned in the face of spurious correlations, how many training examples are necessary, etc.? While Fodor does not focus on acquisition in his criteria, other theories exist which make empirical predictions about how and when specific conceptual representations develop in humans \citep{spelke2007core,carey2009origin}. Future work could translate such predictions into additional unit tests (e.g., measuring learning curves, processing times, etc), in order to diagnose whether current models' errors should be interpreted as failures vs.\ expected signatures of conceptual learning.    

Second, our proposed tests evaluate a system as a whole. Thus, our ability to make claims about neural networks as an implementation of conceptual reasoning is dependent on the quality of the tools available for inspecting neural networks' internals. A particularly fruitful area for future work is finding alternative implementations of \texttt{ablate}. Recent work by \citet{tucker2021if} and \citet{meng2022locating} could be promising places to start.

Finally, we observe interesting trends about the effect of pretraining on conceptual representations. The models we evaluate share the same architecture but have different pretraining regimes.
Only for \texttt{is\_grounded}, and possibly in our layerwise analysis, was there a clear benefit from pretraining. Our results suggested that the pretrained models had an inductive bias for shape over color, and may show more promise in subsequent studies of causality. On other tests, pretraining did not translate to a clear improvement in conceptual structure.

\section{Related Work}
\label{sec:related-work}

Our study follows work on distributional models of semantics, which seeks to interpret computational models based on vectors and neural networks in terms of linguistic and cognitive theories \citep{erk2012vector,lenci2018distributional,boleda2020distributional}. However, we do not take a stand on how vector spaces compare to symbols as models of human language/cognition \textit{at the computational level}. Rather, our study assumes that one prefers a symbolic model at the computational level, and asks whether neural networks could serve as the implementation of such a model.

Closely related is recent work which seeks to answer whether neural networks exhibit properties such as systematicity and compositionality both in NLP \citep{lake2018generalization,yanaka-etal-2019-neural,goodwin-etal-2020-probing,kim-linzen-2020-cogs} and in computer vision \citep{Johnson_2017_CVPR,andreas2016neural}. In contrast to these studies, which assess the final model behavior (analogous to \texttt{predict}), we have additional criteria for how the representations behave (like \texttt{is\_modular}). Also related is prior work which attempts to define mappings between humans' and neural networks' conceptual spaces, e.g., by defining measures of compositionality or groundedness based on how well similarity in vector space reflects similarity according to a symbolic representation \citep{andreas2018measuring,chrupala-alishahi-2019-correlating,merrill2021provable}. Our work differs in that we use a multifaceted suite of evaluation techniques in order to operationalize a specific theory of concepts. 

We use techniques from the broad area of interpretability and analysis of neutral networks. First, work on \textbf{identifying concepts in neural networks} seeks interpretable patterns in the activations and gradients of neural networks, e.g., that unsupervised CNNs encode concepts such as edges \citep{sermanet2013overfeat,le2013building}. Many techniques have been proposed in order to determine which input features are ``important'' to model decisions \citep{ribeiro2016should,sundararajan2017axiomatic, kim2018interpretability, wiegreffe-pinter-2019-attention}.  We employ the method of ``diagnostic classifiers'' \citep{veldhoen2016diagnostic,ettinger-probing,adi2016fine,hupkes2017visualisation}, with the goal of finding high-level concepts which are not directly reducible to input features \citep{kim2018interpretability,tenney2019you}. 
Second, work on \textbf{counterfactual perturbations} attempts to provide causal explanations of model predictions in terms of input features or concepts. Most such work relies on controlled perturbations of the model's input--e.g., manipulating pixels in an image \citep{fong2017interpretable,chang2018explaining,goyal2019counterfactual,goyal2019explaining} or tokens in a string of text \citep{ribeiro-etal-2018-semantically,webster2020measuring,huang-etal-2020-reducing}, though recent methods operate on models' internal representations \citep{vig2020investigating,ravfogel2020null,tucker2021if,meng2022locating}. We employ both types of counterfactual manipulations (we manipulate inputs in \cref{sec:grounded} and representations in \cref{sec:causal}). Unlike prior work, which often treats these counterfactual manipulations as different measures of the same thing, we connect each evaluation to a different aspect of Fodor's theory of concepts.
Finally, our work uses the idea of \textbf{unit testing for neural networks} \citep{adebayo2020debugging,ribeiro-etal-2020-beyond}. 

\section{Conclusion}
We introduce a specification for symbolic conceptual reasoning based on Fodor's theory of concepts. We find evidence that current neural network models are consistent with many predictions of this theory but don't demonstrate a causal connection between the representations of constituent concepts and those of composite concepts. Further investigation into methods for manipulating models' internal representations  may illuminate whether this inconsistency is fundamental to neural networks, or rather a limitation of current analysis tools.

\section*{Acknowledgements}
We would like to thank Roman Feiman, Carsten Eickhoff, Gabor Brody, and Jack Merullo for the helpful discussions, as well as the Brown LUNAR lab. Furthermore, we want to thank our reviewers for their thorough feedback, helping us better present our work. This research was conducted using computational resources and services at the Center for Computation and Visualization, Brown University. This research was supported by the DARPA GAILA program.

\bibliography{concepts}
\bibliographystyle{acl_natbib}
\end{document}